\documentclass[11pt]{article}

\usepackage[margin=1in]{geometry}
\usepackage{times}

\usepackage{amsmath,amsfonts,bm}

\def\eqref#1{equation~\ref{#1}}

\def\1{\bm{1}}

\DeclareMathAlphabet{\mathsfit}{\encodingdefault}{\sfdefault}{m}{sl}
\SetMathAlphabet{\mathsfit}{bold}{\encodingdefault}{\sfdefault}{bx}{n}

\usepackage{amssymb}
\usepackage{algorithm}
\usepackage{algpseudocode}
\usepackage{booktabs}
\usepackage{colortbl}
\usepackage{float}
\usepackage{placeins}
\usepackage{graphicx}
\usepackage[authoryear,round]{natbib}
\usepackage{microtype}
\usepackage{tikz}
\usepackage[most]{tcolorbox}
\usepackage{xcolor}
\usepackage{url}
\usepackage{hyperref}
\usetikzlibrary{arrows.meta,positioning}

\definecolor{skillgradgray}{RGB}{194,198,203}
\definecolor{skillgradedge}{RGB}{122,128,135}
\definecolor{skilloptgray}{RGB}{139,147,156}
\definecolor{skilloptedge}{RGB}{82,91,101}
\definecolor{sspeblue}{RGB}{76,108,146}
\definecolor{sspeedge}{RGB}{48,76,108}
\definecolor{improvementgreen}{RGB}{0,138,76}
\definecolor{degradationred}{RGB}{210,45,45}
\definecolor{ssperowbg}{RGB}{238,238,252}
\definecolor{takeawayaccent}{RGB}{67,96,132}
\definecolor{takeawaybg}{RGB}{245,247,251}
\definecolor{linkblue}{RGB}{32,76,132}

\hypersetup{
    colorlinks=true,
    linkcolor=linkblue,
    citecolor=improvementgreen,
    urlcolor=linkblue,
    pdftitle={Semantic Projection for Continual Self-Evolution of Language Agents},
    pdfauthor={Ziyu Liu, Jun Chen, Lixu Wang}
}

\algrenewcommand{\algorithmiccomment}[1]{%
    \hfill$\triangleright$~{\color{blue!65!black}\scriptsize\itshape #1}}

\newcommand{\method}{\textsc{SSPE}}
\newcommand{\gain}[1]{\textcolor{improvementgreen}{\scriptsize\,$+#1$}}
\newcommand{\loss}[1]{\textcolor{degradationred}{\scriptsize\,$-#1$}}
\newcounter{takeaway}
\newtcolorbox{takeawaybox}{
    colback=takeawaybg,
    colframe=takeawayaccent,
    boxrule=0pt,
    leftrule=2.4pt,
    arc=0pt,
    outer arc=0pt,
    boxsep=0pt,
    left=8pt,
    right=8pt,
    top=6pt,
    bottom=6pt,
    before skip=7pt,
    after skip=7pt}
\newcommand{\takeaway}[1]{%
    \refstepcounter{takeaway}%
    \begin{takeawaybox}
        \textcolor{takeawayaccent}{\textbf{Takeaway \thetakeaway.}} #1
    \end{takeawaybox}}

\title{\textbf{Semantic Projection for Continual Self-Evolution of Language Agents}}
\author{%
    Ziyu Liu$^{1}$ \qquad Jun Chen$^{2}$ \qquad Lixu Wang$^{3}$\\[4pt]
    \small $^{1}$University of Pennsylvania \qquad $^{2}$Hark\\
    \small $^{3}$The Chinese University of Hong Kong, Shenzhen}
\date{}

\begin{document}
\maketitle

\begin{abstract}
Language-model agents increasingly rely on persistent natural-language skills to adapt
beyond their frozen model parameters. When a shared skill is repeatedly revised from a
non-stationary, heterogeneous task stream, however, improvements for new tasks can
overwrite procedures needed for earlier ones. In continual learning, Orthogonal
Gradient Descent (OGD) addresses analogous interference by projecting a new-task gradient
onto a subspace that locally preserves prior predictions. Natural-language skill
revisions, however, have neither gradients nor a canonical vector space in which such a
projection can be performed. We introduce \emph{Semantic-Scope Projected Evolution}
(\method), which transfers the functional principle of gradient projection from
parameter space to behavior space. \method{} treats an unconstrained skill revision as
a proposed update, identifies acquired capabilities with which it may interfere, and
uses the observed gains and regressions to construct a compatible revision rather than
merely rejecting the update. This enables one shared skill to evolve across latent and
recurring task contexts without exposing semantic domain identities to the evolution
model. Across controlled
synthetic streams and heterogeneous real-agent benchmarks, \method{} improves final
cross-domain competence and mitigates forgetting relative to strong skill-evolution
baselines. The evolved skill also retains the strongest average performance after
transfer to a different executor model. These results establish semantic projection as
a promising principle for stable and adaptive self-evolution of language agents.
\end{abstract}

\section{Introduction}

Language-model agents increasingly combine reasoning with tools, environments, and
multi-step action~\citep{yao2023react,zhou2024lats}, yet most are deployed with fixed
behavior. A complementary line of work asks agents to improve from their own
trajectories. Verbal reflection, self-feedback, and distilled experience can repair
later attempts~\citep{shinn2023reflexion,madaan2023selfrefine,zhao2024expel}, while
episodic memories and reusable skills support behavior over longer
horizons~\citep{park2023generative,wang2024voyager}. More recent agents organize this
experience into adaptive memories, reasoning strategies, context playbooks, or
explicit skills~\citep{packer2023memgpt,suzgun2025dynamic,xu2025amem,
ouyang2025reasoningbank,zhang2025ace,yang2026autoskill}. Natural-language skills are a
particularly useful evolution target. They encode procedural knowledge in an
inspectable artifact, can be revised without changing the foundation model, and can be
reused across tasks and even executor models.

The central challenge changes when one shared skill evolves through a heterogeneous
stream. Existing textual optimizers can diagnose traces, aggregate recurring feedback,
validate edits, prune redundant instructions, or maintain optimizer-like
state~\citep{yuksekgonul2025textgrad,wang2026skillgrad,yang2026skillopt,mi2026skillpro}. These mechanisms make individual
updates more systematic, but most are developed primarily within one task
distribution. In deployment, an agent may move among spreadsheet manipulation,
retrieval, function calling, and document understanding. A revision that repairs the
current failures can overwrite, contradict, or deprioritize procedures needed later
when an earlier capability returns. StreamBench and AgentStream show why evolution
must be studied over stateful streams rather than independent examples, and that its
benefit depends strongly on the model, update mechanism, and stream
organization~\citep{wu2024streambench,wei2025evomemory,yan2026agentstream}. We call the
resulting failure \emph{procedural catastrophic forgetting}. Current-domain validation
may reject a locally poor edit, but it cannot expose regressions on absent capabilities
or explain how to preserve a useful new behavior while repairing the damaged one.

This tension parallels continual learning, where a learner must remain plastic on new
tasks without erasing earlier competence~\citep{parisi2019review,delange2022survey}.
Prior work regularizes important parameters, replays representative experience,
allocates task-specific capacity, or constrains new gradients using historical
information~\citep{kirkpatrick2017ewc,zenke2017si,aljundi2018mas,
rolnick2019replay,buzzega2020der,rusu2016progressive,serra2018hat,
lopezpaz2017gem,chaudhry2019agem}. Orthogonal Gradient Descent (OGD) is especially
suggestive. It removes from a new-task gradient the components aligned with directions
to which previous predictions are sensitive, thereby retaining as much useful progress
as possible without changing protected behavior~\citep{farajtabar2020ogd}. A textual
skill, however, has no canonical gradient, subtraction, or behaviorally meaningful
inner product. Embedding-space orthogonality would not imply that two instructions
induce compatible agent behavior. The relevant invariant must therefore be transferred
at the level of observed behavior rather than text geometry.

We introduce \emph{Semantic-Scope Projected Evolution} (\method), a method for
continual skill evolution over heterogeneous task streams. \method{} begins with an
unconstrained revision driven by the current trajectories, predicts which acquired
capabilities the revision may affect, and selectively audits representative historical
behavior. If the proposal causes interference, the model receives both its current
gains and the concrete historical regressions, then iteratively rewrites the proposal
to satisfy the observed constraints or chooses no update. This turns verification from
a terminal accept-or-reject gate into constructive feedback for semantic projection.
Our study formalizes this streaming setting, examines the role of historical capability
evidence through controlled ablations, and evaluates final cross-domain competence,
forgetting, and transfer of evolved skills across executor models. Across controlled
and real heterogeneous streams, the results support using observed historical
conflicts as constructive feedback for repairing an update, rather than using
historical evaluation only as an accept--reject gate.

\section{Related Work}

\paragraph{Persistent experience and self-evolving agents.}
Language agents can learn from interaction by storing reflections, demonstrations,
episodes, or reusable procedures~\citep{shinn2023reflexion,zhao2024expel,
park2023generative,wang2024voyager,packer2023memgpt}. Recent memory systems organize
experience into adaptive notes, linked memories, and retrieved reasoning
strategies~\citep{suzgun2025dynamic,xu2025amem,ouyang2025reasoningbank,
wei2025evomemory}. Surveys consequently treat parameters, prompts, memory, tools,
workflows, and agent architecture as distinct evolution targets~\citep{gao2026survey}.
\method{} focuses on a shared procedural skill and on interference introduced when
that artifact is repeatedly revised across capabilities.

\paragraph{Textual optimization and skill evolution.}
Prompt optimizers search, score, or iteratively refine instructions using model
feedback~\citep{zhou2023ape,yang2024opro,pryzant2023protegi,
fernando2024promptbreeder,khattab2024dspy,agrawal2025gepa}. TextGrad generalizes this
view by propagating natural-language feedback through compound systems~\citep{
yuksekgonul2025textgrad}. Skill-oriented methods add structured diagnosis and semantic
momentum, validation-gated edits, backward pruning, adaptive optimizer state, or
verified procedural units~\citep{wang2026skillgrad,yang2026skillopt,mi2026skillpro}. They improve how updates are
generated or selected. Our concern is complementary: repairing an update after its
useful current behavior and its historical interference have both been observed.

\paragraph{Streaming evaluation and continual learning.}
StreamBench, Evo-Memory, and AgentStream evaluate agents through stateful feedback
sequences rather than isolated examples~\citep{wu2024streambench,wei2025evomemory,
yan2026agentstream}. Continual learning addresses related interference through
regularization~\citep{li2016lwf,kirkpatrick2017ewc,zenke2017si,aljundi2018mas},
replay~\citep{rebuffi2017icarl,rolnick2019replay,buzzega2020der}, architectural
isolation~\citep{rusu2016progressive,serra2018hat,mallya2018packnet}, and gradient
constraints or projections~\citep{lopezpaz2017gem,chaudhry2019agem,
farajtabar2020ogd,saha2021gpm,riemer2019mer}. \method{} does not project a numerical
gradient. It transfers the same stability--plasticity principle by predicting relevant
historical capabilities, measuring behavioral regressions, and asking the evolution
model to construct a compatible textual revision.

\section{Continual Skill Evolution over Heterogeneous Streams}
\label{sec:problem}

We consider a frozen language-model agent $\pi_\theta$ equipped with a persistent
natural-language skill $S_t$. Tasks arrive as a nonstationary stream, grouped only
to update the skill after a completed batch
$B_t=\{x_{t,i}\}_{i=1}^{n_t}$. For each task, the current skill remains fixed while
the agent produces a trajectory and receives terminal feedback:
\begin{equation*}
    \tau_{t,i}\sim \pi_\theta(\,\cdot\mid x_{t,i},S_t),
    \qquad
    r_{t,i}=R(x_{t,i},\tau_{t,i}).
\end{equation*}
Only after all trajectories in $B_t$ terminate may an external evolution procedure
revise the skill,
\begin{equation*}
    S_{t+1}=U(S_t,B_t,\mathcal F_t),
\end{equation*}
where $\mathcal F_t$ may contain trajectories, scores, and textual diagnoses. The
base-model parameters $\theta$ remain frozen. The formulation does not assume that
the stream is partitioned into observed domains: the updater need not receive
semantic domain names, a domain count, or marked distribution changes.

For analysis, we regard the stream as arising from changing mixtures over a set of
latent capability contexts $\mathcal C$. These contexts are evaluator-side constructs,
not inputs to the agent. For $c\in\mathcal C$, let
\begin{equation*}
    J_c(S)
    =
    \mathbb E_{\substack{x\sim P_c\\
                        \tau\sim\pi_\theta(\,\cdot\mid x,S)}}
    \left[R(x,\tau)\right],
\end{equation*}
where $P_c$ is the evaluator-side distribution of tasks requiring capability $c$.
For a candidate skill $S'$, its effect on capability $c$ relative to the current
skill is
\begin{equation*}
    \Delta_c(S';S_t)=J_c(S')-J_c(S_t).
\end{equation*}
A locally useful revision can improve the capabilities represented in $B_t$ while
making $\Delta_c(S';S_t)<0$ for a capability acquired earlier. We call this
\emph{procedural interference}. Because the skill is shared across the entire
stream, such losses persist into later tasks and accumulate as procedural
forgetting. The central problem is therefore to retain improvements supported by
current feedback while preventing regressions on relevant historical capabilities.
This motivates the behavioral analogue of OGD's projection principle developed next.

\section{Semantic-Scope Projected Evolution}
\label{sec:method}

\subsection{From Orthogonal Gradient Descent to Behavioral Projection}

Orthogonal Gradient Descent (OGD) addresses continual learning in a parameterized
model $f(x;w)$ when tasks $\mathcal T_1,\mathcal T_2,\ldots$ arrive sequentially and
earlier training data may no longer be available~\citep{farajtabar2020ogd}. At task
$t$, ordinary gradient descent follows
$g_t=\nabla_w\mathcal L_t(w)$, which is determined by the current data alone. Such a
step can lower the current loss while moving parameters along directions to which
earlier predictions are highly sensitive.

Specifically, for an input $x_{\ell,j}$ from an
earlier task $\ell<t$ and output coordinate $k$, the vector
$u_{\ell,j,k}=\nabla_w f_k(x_{\ell,j};w_\ell^\star)$ describes the local parameter
direction that most changes that output at the checkpoint $w_\ell^\star$ where the
task was learned. OGD retains a selected collection of these
\emph{model-output gradients}
and orthonormalizes them. Let the columns of $U_{<t}$ span the resulting historical
sensitivity subspace. The current gradient is then replaced by its component in the
orthogonal complement:
\begin{equation*}
    \widetilde g_t
    =
    \left(I-U_{<t}U_{<t}^{\top}\right)g_t,
\end{equation*}
and the parameters are updated as $w^+=w-\eta\widetilde g_t$. If a stored
sensitivity $u_{\ell,j,k}$ remains a valid local approximation at $w$, then
\begin{equation*}
    f_k(x_{\ell,j};w^+)-f_k(x_{\ell,j};w)
    \approx
    -\eta u_{\ell,j,k}^{\top}\widetilde g_t
    =0.
\end{equation*}
Thus orthogonality has a behavioral interpretation: the projected step leaves the
protected outputs unchanged to first order. At the same time,
$g_t^{\top}(-\widetilde g_t)=-\lVert\widetilde g_t\rVert_2^2\leq0$, so a nonzero
$-\widetilde g_t$ remains a descent direction for the current loss. OGD therefore
does more than detect interference. It transforms the current update to retain as
much compatible progress as the local parameter geometry permits. If the projection
vanishes, taking no step is the compatible outcome.

OGD stores output rather than loss gradients because the latter may vanish once an
earlier example is fitted, while output gradients still describe sensitivity of the
learned prediction. This yields the principle we transfer: preserve a useful current
update while removing the part that changes protected behavior. A literal projection
is unavailable for textual skills---skill revisions have no canonical subtraction,
inner product, or behaviorally meaningful Euclidean geometry, and embedding
orthogonality does not imply unchanged agent behavior. SSPE therefore projects by
measuring and repairing behavioral conflicts rather than by manipulating text vectors.

\subsection{Design Principle: Projecting Skill Behavior}

SSPE treats projection as conflict-conditioned repair in behavior space.
Table~\ref{tab:ogd-sspe-map} summarizes its analogy to OGD\@. The unconstrained skill revision
is the update favored by current feedback. Capability memory identifies historical
behavior that may be affected, and small representative audits reveal whether the
candidate actually changes that behavior. If they expose a conflict, the evolution
model revises the candidate using both its current-task gains and the measured
historical regressions. If no compatible revision is found, SSPE retains the parent
skill.

\begin{table}[H]
    \centering
    \small
    \caption{SSPE transfers the operational invariant of OGD, not its vector
    operations.}
    \label{tab:ogd-sspe-map}
    \begin{tabular}{p{0.36\columnwidth}p{0.54\columnwidth}}
        \toprule
        \textbf{OGD} & \textbf{SSPE} \\
        \midrule
        Current gradient $g_t$ & Unconstrained revision $\widehat S_t$ \\
        Protected output sensitivities & Capability capsules and audit tasks \\
        Alignment with a protected direction & Measured historical regression \\
        Orthogonal projection & Revision conditioned on observed conflicts \\
        Zero projected step & Explicit no update \\
        \bottomrule
    \end{tabular}
\end{table}

\begin{figure}[t]
    \centering
    \IfFileExists{figures/sspe_framework_v13.png}{%
        \includegraphics[width=0.95\textwidth]{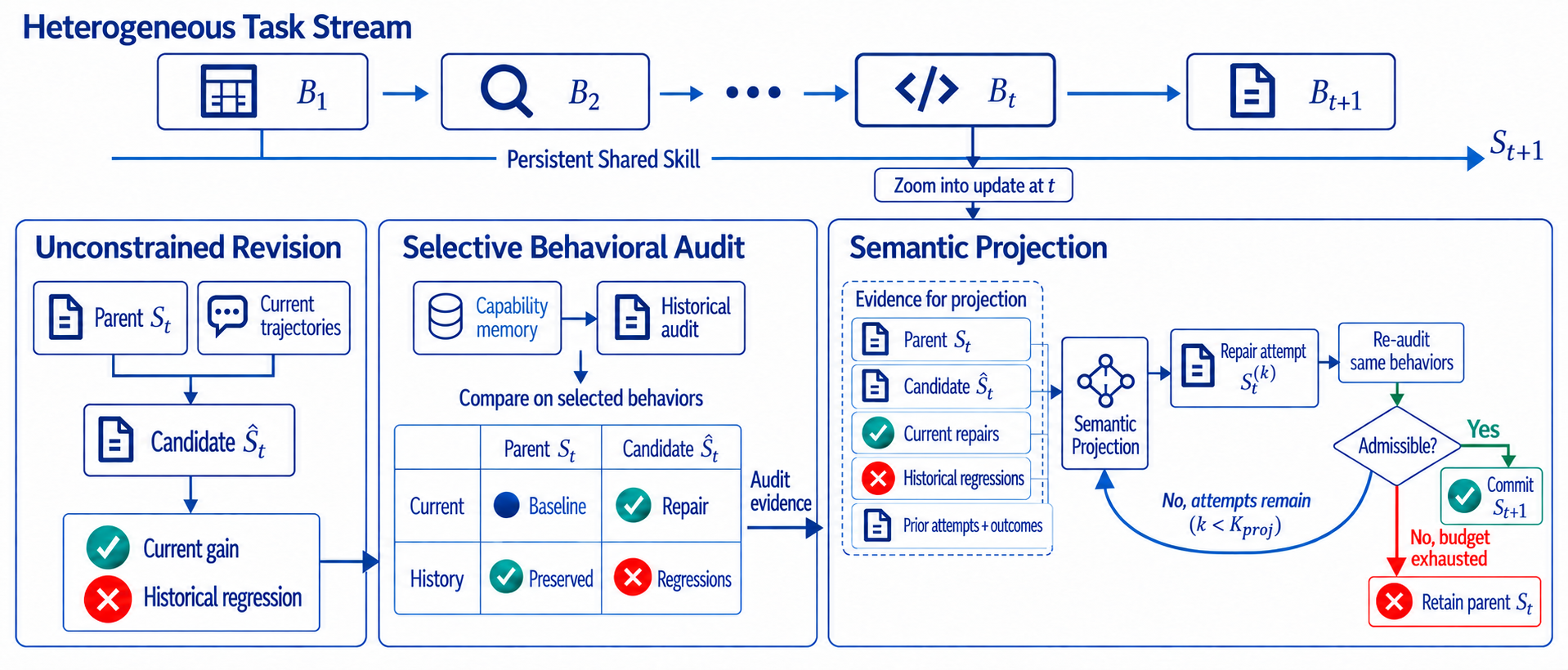}%
    }{%
        \fbox{\parbox[c][2.2in][c]{0.9\textwidth}{\centering Framework figure available in the arXiv source package.}}%
    }
    \caption{SSPE expands one update in a heterogeneous stream. Current evidence
    produces an unconstrained revision; capability memory routes a small audit that
    reveals current gains and historical regressions. An inadmissible candidate is
    repaired and re-audited for at most $K_{\mathrm{proj}}$ rounds. The first
    admissible revision is committed; otherwise, the parent skill is retained.}
    \label{fig:sspe-framework}
\end{figure}

The procedure follows three principles. First, it begins with a \emph{plastic}
proposal, i.e., historical constraints do not restrict how the current evidence may rewrite
the skill. Second, it turns predicted scope into \emph{measured} constraints through
selective audits of earlier capabilities. Third, it uses a failed audit as repair
feedback rather than only as a rejection signal. The model may rewrite the candidate
freely, while the host is responsible only for evidence provenance and the empirical
acceptance test. Figure~\ref{fig:sspe-framework} shows this loop over a heterogeneous
task stream.

\subsection{Capability Evidence and Risk-Directed Audits}

In particular, at update step $t$, SSPE maintains the shared skill $S_t$ and capability memory
$\mathcal M_{t-1}$. After executing $B_t$, a memory operator $\Gamma$ records an
immutable \emph{micro-capsule} and relates it to the accumulated memory:
\begin{equation*}
    (\mathcal M_t,Z_t)
    =\Gamma(\mathcal M_{t-1};B_t,Y_t,D_t).
\end{equation*}
Here, $Y_t$ contains the scored trajectories, $D_t$ their diagnoses, and $Z_t$ is the
set of capability capsules associated with the current batch. A micro-capsule records
task characteristics, observed
successes and failures, scope, contradictions, and uncertainty. Related
micro-capsules form a capability capsule, but the underlying evidence remains
immutable. Each confirmed capsule $h$ points to an opaque audit bank $V_h$ of
representative tasks. The memory summarizes prior behavior; it neither proposes text
nor edits the skill.

SSPE forms the current-task revision before exposing historical constraints. From the
current trajectories, diagnoses, and semantic momentum $\mu_t$, which is a running summary
of recurring diagnoses, a revision operator $\mathcal P$ produces
\begin{equation*}
    (\widehat S_t,\widehat\sigma_t)
    =\mathcal P(S_t;B_t,Y_t,D_t,\mu_t).
\end{equation*}
$\widehat S_t$ is a complete, unconstrained skill proposal, and
$\widehat\sigma_t$ forecasts its intended repairs, affected capabilities, excluded
scope, and uncertainty. The forecast is used for routing, not accepted as evidence
that the revision is safe.

Let $\mathcal H_t$ contain the confirmed historical capsules not in $Z_t$. For every
$h\in\mathcal H_t$, a risk operator $\mathcal R$ compares the proposal and its scope
forecast with the capsule:
\begin{equation*}
    \rho_t(h)=\mathcal R(\widehat S_t,\widehat\sigma_t;\mathcal M_t,h).
\end{equation*}
The review reports qualitative interference risk and uncertainty. SSPE audits the
highest-risk or most-uncertain capability and, when available, one seeded-random
capability predicted safe. The random audit probes false negatives in the semantic
review. An audit router $\mathcal S$ returns the selected historical capabilities
$\mathcal A_t$ and their associated banks:
\begin{align}
    \mathcal A_t
    &=\mathcal S(\rho_t;\mathcal H_t),\nonumber\\
    \mathbb V_t
    &=\{V_t^{\mathrm{cur}}\}
     \cup\{V_h:h\in\mathcal A_t\},
    \label{eq:sspe-bank-suite}
\end{align}
where $V_t^{\mathrm{cur}}$ is a small current-capability comparison bank distinct from
the one-pass stream batch $B_t$. The parent and every candidate are evaluated on this
same suite. Parent outcomes are cached, and only scored behavior and execution traces
are returned to the evolution model.

\subsection{Iterative Projection and Skill Update}

For an audit bank $V$, define its hard-first utility as
\begin{equation*}
    U_V(S)=\bigl(H_V(S),Q_V(S)\bigr),
\end{equation*}
where $H_V$ is hard accuracy and $Q_V\in[0,1]$ is a benchmark-specific secondary
score used only to break hard-score ties. A candidate is admissible when it improves
the current bank lexicographically and causes no net hard loss on any selected
historical bank:
\begin{equation*}
    \mathcal K_t
    =\left\{S'\ \middle|\
    \begin{aligned}
        &U_{V_t^{\mathrm{cur}}}(S')
          \succ_{\mathrm{lex}} U_{V_t^{\mathrm{cur}}}(S_t),\\[-1pt]
        &H_{V_h}(S')\geq H_{V_h}(S_t)
          \quad \forall h\in\mathcal A_t
    \end{aligned}\right\}.
\end{equation*}
This criterion is determined by paired execution, not by the risk prediction.

Set $S_t^{(0)}=\widehat S_t$. Evaluating candidate $S_t^{(k)}$ produces
$E_t^{(k)}$, which contains the parent--candidate utilities, paired repairs and
regressions, and relevant traces for every $V\in\mathbb V_t$. If the candidate is
inadmissible, SSPE collects the evidence observed so far,
\begin{equation}
    \Omega_t^{(k)}
    =\left(\widehat\sigma_t,\rho_t,
      \{(S_t^{(j)},E_t^{(j)})\}_{j=0}^{k}\right),
    \label{eq:sspe-cumulative-feedback}
\end{equation}
and constructs the next candidate by semantic projection:
\begin{equation*}
    S_t^{(k+1)}
    =\Pi_{\mathrm{sem}}
      (S_t^{(k)};S_t,\Omega_t^{(k)},\mathcal M_t),
    \qquad 0\leq k<K_{\mathrm{proj}}.
\end{equation*}
$\Pi_{\mathrm{sem}}$ is an LLM reasoning operator, instead of a closed-form projection operator. It
may rewrite any part of the skill or decline to produce a further revision. Crucially,
it observes both the behavior that the candidate repaired and the behavior that it
broke. The host imposes no repair grammar; it only re-evaluates each candidate on the
unchanged suite $\mathbb V_t$.

The loop stops at the first admissible candidate. Let $\mathcal I_t$ collect the
admissible candidate indices. If it is empty, the skill is not updated:
\begin{equation*}
    \mathcal I_t=\{k\in\{0,\ldots,K_{\mathrm{proj}}\}:
                         S_t^{(k)}\in\mathcal K_t\},
    \qquad
    S_{t+1}=\begin{cases}
        S_t^{(\min\mathcal I_t)}, & \mathcal I_t\neq\varnothing,\\
        S_t, & \text{otherwise.}
    \end{cases}
\end{equation*}
Thus $K_{\mathrm{proj}}$ counts semantic repair attempts after the initial proposal;
$K_{\mathrm{proj}}=1$ recovers the original single-repair implementation. Unlike
validation-only selection, SSPE returns measured conflicts to the optimizer and tests
whether a revised candidate can preserve the current gain. Unlike a fixed repair
rubric, it does not constrain the revision before observing the conflict.

Algorithm~\ref{alg:sspe} assembles the procedure. We use $\mathcal X$ for task
execution, $\mathcal G$ for diagnosis and semantic momentum, $\Gamma$ for capability
memory, $\mathcal P$ for the unconstrained proposal, $\mathcal R$ for risk review,
$\mathcal S$ for audit routing, and $\mathcal V$ for paired audit execution. These
are prompted uses of frozen language models, not trainable task-specific modules. The general procedure receives
neither benchmark identities nor a known domain count. Our controlled evaluation
supplies anonymous consolidation boundaries to isolate projection quality from
context-discovery errors; Appendix~\ref{app:real-stream-protocol} gives the protocol.

\begin{algorithm}[H]
    \caption{Semantic-Scope Projected Evolution}
    \label{alg:sspe}
    \small
    \begin{algorithmic}[1]
        \Require initial skill $S_1$, task batches $\{B_t\}_{t\geq1}$,
                 repair budget $K_{\mathrm{proj}}\geq1$,
                 $\mathcal M_0=\varnothing$, $\mu_0=\varnothing$
        \For{$t=1,2,\ldots$}
            \State $Y_t\gets\mathcal X(B_t;S_t)$
                   \Comment{execute the current skill on the stream}
            \State $(D_t,\mu_t)\gets\mathcal G(B_t,Y_t,\mu_{t-1})$
                   \Comment{extract update evidence}
            \State $(\mathcal M_t,Z_t)\gets
                   \Gamma(\mathcal M_{t-1};B_t,Y_t,D_t)$
                   \Comment{record and consolidate capability evidence}
            \If{$Z_t$ contains no confirmed capability}
                \State $S_{t+1}\gets S_t$; seal $\mathcal M_t$; \textbf{continue}
            \EndIf
            \State $(\widehat S_t,\widehat\sigma_t)\gets
                   \mathcal P(S_t;B_t,Y_t,D_t,\mu_t)$
                   \Comment{form a plastic, unconstrained revision}
            \State $\rho_t(h)\gets
                   \mathcal R(\widehat S_t,\widehat\sigma_t;\mathcal M_t,h)$
                   for each $h\in\mathcal H_t$
            \State $\mathcal A_t\gets\mathcal S(\rho_t;\mathcal H_t)$;
                   form $\mathbb V_t$ by Eq.~\ref{eq:sspe-bank-suite}
                   \Comment{select current and historical checks}
            \State $S_t^{(0)}\gets\widehat S_t$;
                   $E_t^{(0)}\gets
                   \mathcal V(S_t,S_t^{(0)};\mathbb V_t)$
                   \Comment{measure gains and interference}
            \State $k_t^\star\gets\varnothing$
                   \Comment{no compatible candidate found yet}
            \If{$S_t^{(0)}\in\mathcal K_t$}
                \State $k_t^\star\gets0$
                       \Comment{accept the unconstrained revision}
            \Else
                \For{$k=1,\ldots,K_{\mathrm{proj}}$}
                    \State Build $\Omega_t^{(k-1)}$ by
                           Eq.~\ref{eq:sspe-cumulative-feedback}
                           \Comment{accumulate failed attempts}
                    \State $S_t^{(k)}\gets
                           \Pi_{\mathrm{sem}}
                           (S_t^{(k-1)};S_t,\Omega_t^{(k-1)},\mathcal M_t)$
                           \Comment{construct another semantic repair}
                    \State $E_t^{(k)}\gets
                           \mathcal V(S_t,S_t^{(k)};\mathbb V_t)$
                           \Comment{re-audit on identical banks}
                    \If{$S_t^{(k)}\in\mathcal K_t$}
                        \State $k_t^\star\gets k$; \textbf{break}
                               \Comment{accept the first compatible repair}
                    \EndIf
                \EndFor
            \EndIf
            \If{$k_t^\star=\varnothing$}
                \State $S_{t+1}\gets S_t$
                       \Comment{budget exhausted; retain the parent}
            \Else
                \State $S_{t+1}\gets S_t^{(k_t^\star)}$
                       \Comment{commit the accepted revision}
            \EndIf
            \State Seal $S_{t+1}$, $\{E_t^{(k)}\}$, and $\mathcal M_t$
                   \Comment{persist an auditable state}
        \EndFor
    \end{algorithmic}
\end{algorithm}

\section{Experiments}

We organize the evaluation around three questions. First, can selective historical
audits and semantic repair reduce interference in a controlled setting? Second, does
the resulting procedure leave a stronger shared skill after a heterogeneous task
stream? Third, does that skill remain useful when the executor model changes? We
compare SSPE with SkillGrad \citep{wang2026skillgrad}, SkillOpt \citep{yang2026skillopt}, and a fixed initial skill under matched task,
model, and evaluation conditions. Appendix~\ref{app:real-stream-protocol} gives the
complete real-stream protocol and Appendix~\ref{app:synthetic-mechanism} specifies the
controlled simulator.

\subsection{Controlled Synthetic Experiment}

We first isolate cross-capability interference in a provider-free simulator. Each
world represents the agent by latent capability scores, and every proposed revision
improves the active capability while potentially helping or harming earlier ones. All
policies receive paired worlds and environmental random draws; only their update
decisions differ.

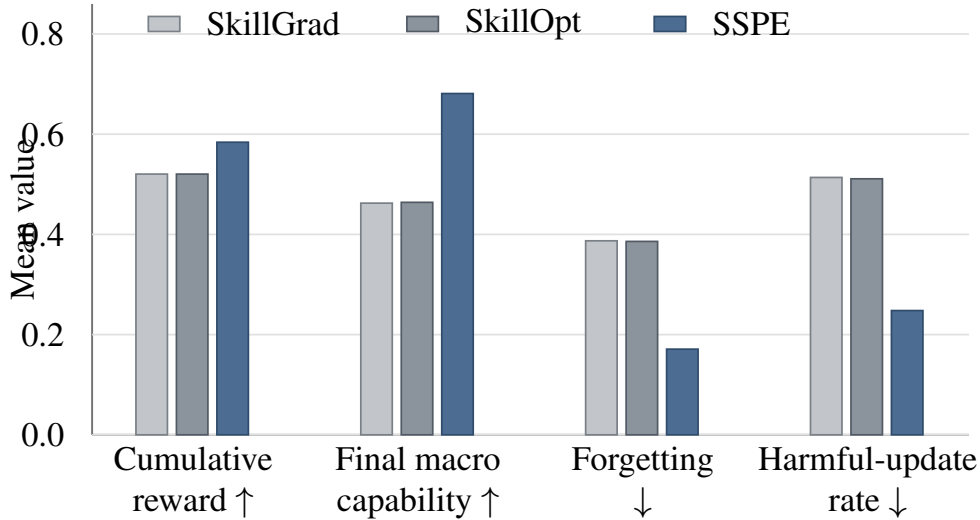
\begin{figure}[H]
    \centering
    \resizebox{0.80\columnwidth}{!}{%
    \begin{tikzpicture}[x=0.92cm,y=4.1cm,font=\scriptsize]
        \draw[black!55,line width=0.45pt] (0.65,0) -- (8.45,0);
        \draw[black!55,line width=0.45pt] (0.65,0) -- (0.65,0.86);
        \foreach \y/\lab in {0/0.0,0.2/0.2,0.4/0.4,0.6/0.6,0.8/0.8} {
            \draw[black!12] (0.65,\y) -- (8.45,\y);
            \node[anchor=east] at (0.55,\y) {\lab};
        }
        \node[rotate=90] at (0.03,0.43) {Mean value};

        \filldraw[fill=skillgradgray,draw=skillgradedge,line width=0.4pt]
            (1.04,0) rectangle (1.32,0.5204);
        \filldraw[fill=skilloptgray,draw=skilloptedge,line width=0.4pt]
            (1.40,0) rectangle (1.68,0.5204);
        \filldraw[fill=sspeblue,draw=sspeedge,line width=0.4pt]
            (1.76,0) rectangle (2.04,0.5840);
        \node[align=center] at (1.55,-0.095) {Cumulative\\reward $\uparrow$};

        \filldraw[fill=skillgradgray,draw=skillgradedge,line width=0.4pt]
            (3.04,0) rectangle (3.32,0.4624);
        \filldraw[fill=skilloptgray,draw=skilloptedge,line width=0.4pt]
            (3.40,0) rectangle (3.68,0.4640);
        \filldraw[fill=sspeblue,draw=sspeedge,line width=0.4pt]
            (3.76,0) rectangle (4.04,0.6812);
        \node[align=center] at (3.55,-0.095) {Final macro\\capability $\uparrow$};

        \filldraw[fill=skillgradgray,draw=skillgradedge,line width=0.4pt]
            (5.04,0) rectangle (5.32,0.3872);
        \filldraw[fill=skilloptgray,draw=skilloptedge,line width=0.4pt]
            (5.40,0) rectangle (5.68,0.3860);
        \filldraw[fill=sspeblue,draw=sspeedge,line width=0.4pt]
            (5.76,0) rectangle (6.04,0.1710);
        \node[align=center] at (5.55,-0.095) {Forgetting\\$\downarrow$};

        \filldraw[fill=skillgradgray,draw=skillgradedge,line width=0.4pt]
            (7.04,0) rectangle (7.32,0.5137);
        \filldraw[fill=skilloptgray,draw=skilloptedge,line width=0.4pt]
            (7.40,0) rectangle (7.68,0.5110);
        \filldraw[fill=sspeblue,draw=sspeedge,line width=0.4pt]
            (7.76,0) rectangle (8.04,0.2480);
        \node[align=center] at (7.55,-0.095) {Harmful-update\\rate $\downarrow$};

        \filldraw[fill=skillgradgray,draw=skillgradedge,line width=0.4pt]
            (1.15,0.805) rectangle (1.43,0.835);
        \node[anchor=west] at (1.52,0.820) {SkillGrad};
        \filldraw[fill=skilloptgray,draw=skilloptedge,line width=0.4pt]
            (3.45,0.805) rectangle (3.73,0.835);
        \node[anchor=west] at (3.82,0.820) {SkillOpt};
        \filldraw[fill=sspeblue,draw=sspeedge,line width=0.4pt]
            (5.65,0.805) rectangle (5.93,0.835);
        \node[anchor=west] at (6.02,0.820) {SSPE};
    \end{tikzpicture}%
    }
    \caption{Absolute outcomes of SkillGrad, SkillOpt and SSPE in the complete controlled sweep. Arrows indicate the preferred direction.}
    \label{fig:synthetic-results}
\end{figure}

Figure~\ref{fig:synthetic-results} shows that current-only gating leaves SkillOpt close
to SkillGrad. SSPE instead produces substantially less forgetting and fewer harmful
updates, yielding stronger cumulative and final performance. This pattern indicates
that selective historical observation changes the long-run outcome of otherwise
identical proposals: SSPE gives up a small amount of immediate plasticity while
retaining substantially more capability over the stream. The appendix reports the
complete stability--plasticity trade-off.

\subsection{Practical Heterogeneous Streams}

We next evolve one shared skill through
SpreadsheetBench $\rightarrow$ SearchQA $\rightarrow$ BFCL $\rightarrow$ DocVQA\@.
Each benchmark contributes 40 tasks, processed in batches of eight. All methods start
from the same neutral skill and use GPT-5.4 for both task execution and skill
evolution. We compare Static skill, SkillGrad, SkillOpt, and SSPE\@. Static skill is
the shared general-purpose initialization used by all evolving methods, held fixed
throughout the stream. After evolution, we freeze each skill and evaluate it on 80
unseen tasks from every benchmark. The frozen-skill evaluation measures what the
shared skill retains across domains. The one-pass stream, verification banks, held-out
partitions, baseline configurations, and benchmark-specific metrics are detailed in
Appendix~\ref{app:real-stream-protocol}.

\begin{table}[H]
    \centering
    \small
    \caption{Accuracy (\%) on held-out tasks for the no-skill executor and final
    frozen skills. Colored suffixes show change from the no-skill executor.}
    \label{tab:heldout-results}
    \setlength{\tabcolsep}{2.7pt}
    \begin{tabular}{lccccc}
        \toprule
        Method & SpreadsheetBench & SearchQA & BFCL & DocVQA & Average \\
        \midrule
        No skill & \textbf{78.75} & 73.75 & 45.00 & 90.00 & 71.88 \\
        Static skill & 75.00\loss{3.75} & 76.25\gain{2.50} & 47.50\gain{2.50} & 91.25\gain{1.25} & 72.50\gain{0.62} \\
        SkillGrad & 72.50\loss{6.25} & 76.25\gain{2.50} & 55.00\gain{10.00} & 92.50\gain{2.50} & 74.06\gain{2.18} \\
        SkillOpt & 73.75\loss{5.00} & \textbf{77.50}\gain{3.75} & 52.50\gain{7.50} & 88.75\loss{1.25} & 73.13\gain{1.25} \\
        \rowcolor{ssperowbg}
        \method{} & 73.75\loss{5.00} & 76.25\gain{2.50} & \textbf{65.00}\gain{20.00} & \textbf{95.00}\gain{5.00} & \textbf{77.50}\gain{5.62} \\
        \bottomrule
    \end{tabular}
\end{table}

\paragraph{SSPE leaves the strongest final skill.}
Table~\ref{tab:heldout-results} shows that SSPE leaves the strongest final skill on
average. Its clearest advantage appears on BFCL, consistent with the hypothesis that
semantic projection is useful when tool-use procedures must remain compatible; it also
matches the best result on DocVQA\@. The no-skill executor remains strongest on
SpreadsheetBench, while SkillOpt
is strongest on SearchQA, so the improvement is not a uniform shift on every benchmark.
Instead, the whole-method comparison shows that SSPE combines the benchmark-specific
procedures more successfully overall.

\takeaway{Semantic projection leaves a stronger cross-domain skill. The advantage is
largest where procedural compatibility matters most, rather than appearing as a
uniform gain on every benchmark.}

\subsection{Cross-Model Generalization}

We replace the GPT-5.4 executor with GPT-4.1 while keeping every evolved skill fixed.
No further evolution or adaptation is allowed, and every skill is evaluated on the
same held-out tasks as above. This experiment tests whether the learned procedures are
portable instructions or merely exploit behavior specific to the model that produced
them.

\begin{table}[H]
    \centering
    \small
    \caption{Cross-model transfer of GPT-5.4-evolved skills to GPT-4.1 on the same
    held-out tasks. Values are accuracy (\%); colored suffixes show change from the
    no-skill executor.}
    \label{tab:gpt41-transfer}
    \setlength{\tabcolsep}{2.7pt}
    \begin{tabular}{lccccc}
        \toprule
        Frozen skill & SpreadsheetBench & SearchQA & BFCL & DocVQA & Average \\
        \midrule
        No skill & \textbf{50.00} & 61.25 & 57.50 & 72.50 & 60.31 \\
        Static skill & 38.75\loss{11.25} & 61.25\gain{0.00} & 57.50\gain{0.00} & 75.00\gain{2.50} & 58.13\loss{2.18} \\
        SkillGrad & 41.25\loss{8.75} & \textbf{65.00}\gain{3.75} & 60.00\gain{2.50} & 81.25\gain{8.75} & 61.88\gain{1.57} \\
        SkillOpt & 36.25\loss{13.75} & 61.25\gain{0.00} & 56.25\loss{1.25} & 72.50\gain{0.00} & 56.56\loss{3.75} \\
        \rowcolor{ssperowbg}
        \method{} & 47.50\loss{2.50} & 57.50\loss{3.75} & \textbf{66.25}\gain{8.75} & \textbf{83.75}\gain{11.25} & \textbf{63.75}\gain{3.44} \\
        \bottomrule
    \end{tabular}
\end{table}

\paragraph{SSPE remains strongest after cross-model transfer.}
Table~\ref{tab:gpt41-transfer} shows that SSPE remains the strongest method on average
when GPT-4.1 executes skills evolved by GPT-5.4. The transferred SSPE skill is
particularly effective on BFCL and DocVQA, the same domains in which structured
procedures and output contracts are central. No skill is strongest on SpreadsheetBench
and SkillGrad is strongest on SearchQA, so the result is not universal across task
types. Nevertheless, retaining the best average after replacing the executor supports
the interpretation that SSPE learns portable procedural guidance rather than exploiting
idiosyncrasies of the evolution model.

\takeaway{SSPE retains the strongest average accuracy after the executor model changes.
This suggests that semantic projection produces reusable procedural guidance rather
than a skill tailored only to the model that evolved it.}

\subsection{Ablations: Historical Evidence and Projection Depth}

We examine two design choices. \emph{SSPE-NoHistory} retains the same proposal,
current-task verification, and repair machinery as SSPE, but removes historical
capability capsules and audits. We also vary the maximum number of semantic projection
attempts after the unconstrained proposal. The default uses
$K_{\mathrm{proj}}=1$; the completed alternatives use $K_{\mathrm{proj}}=2$ and $5$
with the same stream, initial skill, audit banks, acceptance rule, and GPT-5.4 model.

\begin{table}[H]
    \centering
    \small
    \caption{Held-out accuracy (\%) when removing historical evidence or changing
    semantic projection depth.}
    \label{tab:sspe-ablation}
    \begin{tabular}{lcrrrrr}
        \toprule
        Variant & $K_{\mathrm{proj}}$ & SpreadsheetBench & SearchQA & BFCL & DocVQA & Average \\
        \midrule
        SSPE-NoHistory & 1 & \textbf{75.00} & 73.75 & 50.00 & \textbf{95.00} & 73.44 \\
        \rowcolor{ssperowbg}
        \method{} (default) & 1 & 73.75 & \textbf{76.25} & \textbf{65.00} & \textbf{95.00} & \textbf{77.50} \\
        \method{} & 2 & \textbf{76.25} & 73.75 & 50.00 & 92.50 & 73.13 \\
        \method{} & 5 & 71.25 & 73.75 & 46.25 & 90.00 & 70.31 \\
        \bottomrule
    \end{tabular}
\end{table}

Table~\ref{tab:sspe-ablation} yields two complementary observations. First, removing
historical evidence eliminates the overall advantage in this study, even though the
proposal and current-task repair machinery remain unchanged. This indicates that
historical behavioral evidence is important for cross-domain compatibility. Second,
allowing more repair attempts does not improve the current system. The default
$K_{\mathrm{proj}}=1$ is strongest overall, and the gap widens as the budget grows.
Similar evidence is observed for GPT-4.1
(Appendix Table~\ref{tab:projection-depth-gpt41}).
The training traces in Appendix~\ref{app:projection-depth}, particularly
Table~\ref{tab:projection-depth-training}, show that later rounds generate
additional candidates but never an additional compatible update. The result favors
one feedback-complete repair attempt, not repeated unconstrained rewriting after a
conflict has already been exposed.

The budget runs are independent GPT-5.4 optimization lineages because provider
sampling is not seedable. We therefore use them to select the default stopping budget,
not to claim that deeper projection is intrinsically harmful.
Appendix~\ref{app:projection-depth} reports their training behavior and a conservative
sensitivity analysis.

\takeaway{Removing historical evidence eliminates SSPE's aggregate advantage in this
study, despite retaining the proposal and current-task repair machinery. Across both
backbones, increasing the projection budget does not improve final accuracy.}

\section{Limitations}

We note some limitations of our study. First, the current continual streams contain a finite
set of domains whose tasks arrive in groups, whereas practical deployments may involve
an open-ended sequence of domains with less structured transitions. Second, our
experiments are conducted primarily under a fixed task order. Studying SSPE's
robustness across alternative orders, recurrent sequences, and interleaved streams is
an important direction for future work.

\section{Conclusion}

Continual agent self-evolution requires more than improving the most recently observed
tasks, because a locally useful skill revision can silently damage procedures needed
elsewhere in a heterogeneous stream. SSPE transfers the functional principle of
gradient projection to behavioral space. It begins with a plastic current-task
proposal, uses semantic scope reasoning and selective audits to expose behavioral
conflicts, and semantically projects the proposal into a revised skill that preserves
its useful behavior while satisfying the observed historical constraints. Across
controlled and real heterogeneous streams, SSPE produces a stronger final cross-domain
skill than current-only evolution and validation-gated baselines, and the resulting
skill remains strongest on average after transfer to a different executor model. The
ablation shows that historical evidence supplies necessary constraints, but semantic
projection is what makes those constraints constructive: rather than merely rejecting
an interfering update, SSPE uses the observed conflict to build a compatible one.

\subsection*{AI Use Statement}

In this work, generative AI tools were used to assist with literature discovery,
polishing the manuscript, modifying scientific figures, and formatting the manuscript
and references. GPT models were also used as experimental components of SSPE and the
comparison baselines, as documented in the experimental setup. The authors checked
suggested references against original papers and official records, tested AI-assisted
code and debugging changes, inspected the resulting figures and formatting, and
verified reported results against experiment artifacts. The authors reviewed all
AI-assisted material and take responsibility for the final text, claims, code, and
artifacts.

\subsection*{Reproducibility Statement}

Section~\ref{sec:method} and Algorithm~\ref{alg:sspe} specify SSPE's update rule and
information flow. Appendix~\ref{app:real-stream-protocol} documents the benchmark
interfaces, stream construction, task partitions, initial skill, model settings,
baseline configurations, and frozen-skill evaluation protocol;
Appendix~\ref{app:projection-depth} reports the projection-budget diagnostics;
Appendix~\ref{app:synthetic-mechanism} fully specifies the controlled simulator,
policies, metrics, and sweep; and Appendix~\ref{app:qualitative-example} provides a
complete qualitative skill-and-trajectory example. We also provide an anonymized
supplementary package containing the source code, prompts, configurations,
split and task manifests, initial and evolved skill artifacts, and task-level result
records needed to reproduce the reported experiments.

\bibliography{references}
\bibliographystyle{plainnat}

\clearpage
\appendix
\section{Real-Stream Experimental Protocol}
\label{app:real-stream-protocol}

\paragraph{Benchmarks and stream construction.}

The experiment uses four agent capabilities with distinct interfaces and output
contracts. SpreadsheetBench requires editing and validating workbooks; SearchQA
requires retrieval and exact-answer synthesis; BFCL requires structured function
selection and argument construction; and DocVQA requires extracting answers from
document images. Table~\ref{tab:benchmark-protocol} summarizes how each benchmark is
scored.

\begin{table}[H]
    \centering
    \small
    \caption{Benchmarks and evaluation signals in the heterogeneous stream. Hard
    success is the primary metric in every domain; partial scores are diagnostic.}
    \label{tab:benchmark-protocol}
    \begin{tabular}{lll}
        \toprule
        Benchmark & Hard outcome & Secondary signal \\
        \midrule
        SpreadsheetBench & task success & cell accuracy \\
        SearchQA & exact answer & --- \\
        BFCL & task success & turn-prefix accuracy \\
        DocVQA & exact answer & ANLS \\
        \bottomrule
    \end{tabular}
\end{table}

Each benchmark contributes three disjoint partitions: 40 one-pass stream tasks, an
eight-task verification bank, and 80 held-out tasks. Identities and their order are
fixed before any method is run. Stream feedback becomes available only after the
corresponding trajectory terminates, and a stream identity is never replayed to choose
an update. Verification banks may be queried only by methods whose native update rule
uses them. Held-out identities remain sealed until every final skill is frozen and are
selected by a fixed hash order without consulting any method's outcome.

The primary schedule is
\begin{equation*}
    \text{SpreadsheetBench}
    \rightarrow \text{SearchQA}
    \rightarrow \text{BFCL}
    \rightarrow \text{DocVQA}.
\end{equation*}
Each domain contributes five consecutive batches of eight tasks. To isolate semantic
projection from context-discovery errors, the evaluated SSPE implementation receives
an anonymous consolidation boundary and an opaque verification-bank identifier at
each transition. The evolution model never receives the benchmark name or a semantic
domain label. This is a controlled specialization of the boundary-free formulation
in Section~\ref{sec:problem}.

\paragraph{Shared skill and execution.}

All evolving methods begin from the same neutral natural-language skill. It contains
general instructions for interpreting a task, using available tools, checking the
result, and respecting the requested output contract; it contains no benchmark-specific
solution procedure. We call this artifact the \emph{Static skill} when it is held
fixed throughout the stream. The \emph{No skill} evaluation removes this artifact
entirely and is used only where a matched result is available.

GPT-5.4 is used for task execution and every model-based evolution role. Reasoning
effort is set to \texttt{none}, and provider-default decoding is retained. All methods
share the same task prompts, tools, scorers, batch order, and trajectory limits. An
eight-task batch is executed in two waves with at most four trajectories in flight;
dependent turns within a trajectory and skill updates between batches remain
sequential. The foundation-model parameters are frozen, so only the external skill
and method-specific optimizer state change.

\paragraph{Compared methods and information access.}

Table~\ref{tab:method-protocol} distinguishes the update information available to each
method. SkillGrad receives completed stream trajectories but no verification bank.
SkillOpt uses only the active domain's bank. SSPE-NoHistory shares SSPE's proposer,
current-bank verification, and repair operator, but receives no historical capsules or
historical audits. Full SSPE may inspect the current bank, the historical bank judged
most at risk, and one seeded-random historical bank predicted safe when available.

\begin{table*}[t]
    \centering
    \small
    \caption{Method configurations and information available during evolution.}
    \label{tab:method-protocol}
    \begin{tabular}{p{0.16\textwidth}p{0.25\textwidth}p{0.24\textwidth}p{0.23\textwidth}}
        \toprule
        Method & Update mechanism & Verification access & Persistent optimizer state \\
        \midrule
        Static skill & none & none & none \\
        SkillGrad & diagnosis, semantic momentum, patching & none & semantic momentum \\
        SkillOpt & reflection, bounded editing, validation gate & current bank only & rejected-edit memory \\
        SSPE-NoHistory & free proposal and semantic repair & current bank only & current evidence only \\
        \rowcolor{ssperowbg}
        SSPE & risk-directed semantic projection & current and selected historical banks & capability capsules \\
        \bottomrule
    \end{tabular}
\end{table*}

SkillOpt is run for one pass over the stream with batch size eight, reflection
minibatches of four, merge size two, and a textual edit budget that decays from two to
one. It requires strict improvement on the current verification bank and receives no
historical replay. SkillGrad and SSPE likewise perform one update opportunity after
each completed stream batch. SSPE permits one semantic repair attempt in its default
configuration.

\paragraph{Frozen-skill evaluation.}

After the stream, each final skill is frozen and evaluated once on the same 80 held-out
tasks per benchmark. We report hard accuracy separately for every benchmark and the
equal-domain average. A \emph{repair} is a held-out task solved by SSPE and missed by
the comparison method; a \emph{regression} reverses that relation. Secondary benchmark
scores are retained for diagnostic analyses but do not override hard success.

For cross-model transfer, the GPT-5.4-evolved artifacts are left unchanged while
GPT-4.1 replaces the executor. Task identities, prompts, tools, and scorers are the
same as in the GPT-5.4 held-out evaluation, and no further evolution, adaptation, or
checkpoint selection is allowed.

\section{Projection-Budget Diagnostics}
\label{app:projection-depth}

\paragraph{Protocol.}

$K_{\mathrm{proj}}$ is the maximum number of semantic projection calls made after
the unconstrained proposal. The settings $K_{\mathrm{proj}}\in\{1,2,5\}$ therefore
permit at most two, three, and six evaluated candidates per update, respectively.
The stream, initial skill, verification banks, risk router, acceptance rule, and
held-out tasks are fixed across settings. Because GPT-5.4 sampling is not seedable,
the settings are independent optimization lineages rather than deterministic forks
from identical proposals. The comparison should therefore be read as a stopping-budget
ablation, not as a causal estimate of the value of one additional repair call.

\paragraph{Observed training behavior.}

\begin{table}[H]
    \centering
    \small
    \caption{How the projection budget was used during the 20 stream updates. An
    additional round has index $k\geq2$, beyond the default first repair.}
    \label{tab:projection-depth-training}
    \begin{tabular}{lrrrr}
        \toprule
        $K_{\mathrm{proj}}$ & Batches entering $k\geq2$ & Later repairs accepted & Total commits & Preq. (\%) \\
        \midrule
        1 & --- & --- & 4 & 56.25 \\
        2 & 3 & 0 & 3 & 55.63 \\
        5 & 6 & 0 & 3 & 54.38 \\
        \bottomrule
    \end{tabular}
\end{table}

The $K_{\mathrm{proj}}=5$ lineage exhausts all five repair attempts on five batches. Nevertheless,
neither larger-budget lineage commits a candidate from a round beyond the first.
Table~\ref{tab:sspe-ablation} consequently reports the final frozen-skill result only
once, in the main ablation table. Relative to $K_{\mathrm{proj}}=1$, the
$K_{\mathrm{proj}}=2$ skill yields 12 held-out repairs and 26 regressions, while the
$K_{\mathrm{proj}}=5$ skill yields 8 repairs and 31
regressions. The diagnostic traces therefore attribute the lower aggregate scores to
different final skills, not to a successful late repair that subsequently failed to
generalize.

\paragraph{Conservative failure sensitivity.}

One four-task SpreadsheetBench chunk is conservatively scored as four failures for
both $K_{\mathrm{proj}}=2$ and $K_{\mathrm{proj}}=5$ after ambiguous provider responses. An upper-bound sensitivity
analysis that changes all four outcomes to successes raises their overall accuracies
to 74.38\% and 71.56\%, respectively, leaving both below the default's 77.50\%.

\paragraph{Replication with GPT-4.1.}

We repeat the projection-budget ablation with GPT-4.1 for every agent role, training
three fresh skill lineages from the same initial skill and evaluating them on the
same held-out tasks. This is an independent repetition of the complete evolution
procedure, rather than execution of the GPT-5.4-evolved skills with a different
model. Table~\ref{tab:projection-depth-gpt41} shows the same aggregate ordering as
the GPT-5.4 study: one semantic projection attempt is strongest on average, while
larger budgets provide no improvement.

\begin{table}[H]
    \centering
    \small
    \caption{GPT-4.1 repetition of the projection-budget ablation. Values are
    held-out accuracy (\%) after independently training each lineage with
    $K_{\mathrm{proj}}\in\{1,2,5\}$.}
    \label{tab:projection-depth-gpt41}
    \setlength{\tabcolsep}{3.2pt}
    \begin{tabular}{lccccc}
        \toprule
        $K_{\mathrm{proj}}$ & SpreadsheetBench & SearchQA & BFCL & DocVQA & Average \\
        \midrule
        \rowcolor{ssperowbg}
        1 & \textbf{48.75} & 65.00 & \textbf{60.00} & 87.50 & \textbf{65.31} \\
        2 & 46.25 & \textbf{66.25} & 53.75 & \textbf{88.75} & 63.75 \\
        5 & 41.25 & 61.25 & 57.50 & 72.50 & 58.13 \\
        \bottomrule
    \end{tabular}
\end{table}

The $K_{\mathrm{proj}}=2$ and $K_{\mathrm{proj}}=5$ lineages trail the default by
1.56 and 7.19 percentage points, respectively. Because all three skills arise from
independent, non-seedable optimization lineages, the repetition does not isolate the
causal effect of an individual extra repair call. It nevertheless shows that the
negative larger-budget result is not unique to GPT-5.4 and supports retaining
$K_{\mathrm{proj}}=1$ as the default across both backbones.

\section{Controlled Synthetic Study Specification}
\label{app:synthetic-mechanism}

\paragraph{Purpose and abstraction level.}

The synthetic study isolates the decision problem created by cross-capability
interference. It is a deterministic, provider-free simulator rather than an agent
benchmark. Although the released manifest contains task-like surfaces inspired by
function calling, spreadsheet manipulation, retrieval QA, and document QA, the policy
simulation does not execute those prompts. It represents the agent directly by a
latent capability vector and represents a proposed skill revision by its behavioral
effects on that vector. Consequently, this study tests the projection mechanism under
controlled interference; it does not reproduce the textual behavior of native
SkillGrad, SkillOpt, or SSPE\@.

\paragraph{World and stream construction.}

A world contains $K$ latent capabilities. Immediately before step $t$, the simulated
agent has capability vector

\begin{equation*}
    \mathbf q_t=(q_{t,1},\ldots,q_{t,K})\in[0,1]^K,
    \qquad q_{0,c}\sim\operatorname{Uniform}(0.40,0.55).
\end{equation*}

Each capability contributes $m$ batches of $b=8$ tasks. In a contiguous stream, all
$m$ batches from one capability arrive before the next capability. In a recurrent
stream, the simulator cycles through all $K$ capabilities before beginning the next
round of batches. If capability $c_t$ is active, its prequential batch score is

\begin{equation}
    a_t=\frac{1}{b}\sum_{i=1}^{b}Y_{t,i},
    \qquad Y_{t,i}\sim\operatorname{Bernoulli}(q_{t,c_t}).
    \label{eq:synthetic-batch-score}
\end{equation}

The simulator then constructs one shared proposed update. Its current-capability gain
is sampled as

\begin{equation*}
    \delta_{t,c_t}\sim\operatorname{Uniform}(0.075,0.145).
\end{equation*}

For every previously encountered capability $c\neq c_t$, the proposal yields positive
transfer with probability $0.25$,

\begin{equation*}
    \delta_{t,c}=\beta Z_c,
    \qquad \beta=0.025,
    \qquad Z_c\sim\operatorname{Uniform}(0.5,1.5),
\end{equation*}

and otherwise yields interference

\begin{equation*}
    \delta_{t,c}=-\lambda Z'_c,
    \qquad Z'_c\sim\operatorname{Uniform}(0.55,1.45).
\end{equation*}

Here $\lambda$ is the configured interference level. Under gradual drift, the
interference magnitude at within-capability batch $j$ is multiplied by
$1+0.4j/(m-1)$. Under abrupt drift, it is multiplied by $1.65$ from the midpoint of a
capability sequence onward. Applied updates are clipped to $[0,1]^K$.

\begin{table}[H]
    \centering
    \small
    \caption{Controlled synthetic sweep. The main factorial uses $K=4$ and $m=5$;
    $K$ and $m$ are additionally varied through one-factor scaling slices.}
    \label{tab:synthetic-sweep}
    \begin{tabular}{ll}
        \toprule
        Quantity & Values \\
        \midrule
        Worlds per configuration & 100 \\
        Latent capabilities $K$ & $2,4,8,16$ \\
        Batches per capability $m$ & $2,5,10$ \\
        Tasks per batch $b$ & 8 \\
        Interference $\lambda$ & $0,0.10,0.22$ \\
        Positive-transfer scale $\beta$ & $0.025$ \\
        Scope-error rate $\epsilon$ & $0.05,0.20,0.40$ \\
        Projection success $p_{\mathrm{proj}}$ & $0.70,0.90$ \\
        Current-gain retention $\rho$ & $0.86$ \\
        Drift & none, gradual, abrupt \\
        Arrival pattern & contiguous, recurrent \\
        \bottomrule
    \end{tabular}
\end{table}

The full factorial crosses interference, scope error, projection success, drift, and
arrival pattern at $K=4,m=5$. Scaling slices vary $K$ and $m$ at
$\lambda=0.22$, $\epsilon=0.20$, $p_{\mathrm{proj}}=0.90$, no drift, and recurrent
arrival. After removing duplicate configurations, the study contains 113
configurations and 11,300 paired worlds. The preregistered nonlinear analysis excludes
$\lambda=0$, leaving 77 configurations and 7,700 paired worlds.

\paragraph{Simulated observation and projection.}

Scope reasoning is represented by a noisy predictor. A genuinely harmed historical
capability is marked risky with probability $1-\epsilon$. A non-harmed capability is
spuriously marked risky with probability $0.35\epsilon$. SSPE checks one predicted-risk
capability when available; otherwise it falls back to a previously observed
capability. It additionally checks one seeded-random capability predicted safe when
one is available. Each selected audit represents eight task executions.

If a checked historical capability is harmed, semantic projection removes its negative
effect with probability $p_{\mathrm{proj}}$. Every successful repair retains a fraction
$\rho=0.86$ of the remaining current-capability gain. If a checked regression remains
after attempted repair, the candidate is rejected and the capability vector remains
unchanged. A candidate must also strictly improve the current capability. This
simulates the information flow of SSPE---predict risk, measure selected behavior, and
repair an observed conflict---without modeling the language used to produce the
revision.

\paragraph{Compared policies.}

\begin{table}[H]
    \centering
    \small
    \caption{Policies in the controlled simulator. SkillGrad and SkillOpt denote
    mechanism-level abstractions of their update decisions, not executions of the
    corresponding released methods.}
    \label{tab:synthetic-policies}
    \begin{tabular}{p{0.18\textwidth}p{0.74\textwidth}}
        \toprule
        Policy & Update rule \\
        \midrule
        Static skill & Retain $\mathbf q_t$ at every step. \\
        SkillGrad & Commit every proposal; perform no current-validation or
        historical-capability check. This is the primary baseline. \\
        SkillOpt & Apply a current-capability improvement gate but perform no
        historical check. Because generated proposals improve the current capability,
        this policy is usually close to SkillGrad. \\
        Random replay & Audit one uniformly sampled previous capability. Attempt repair
        if the audit discovers interference; reject the proposal if checked interference
        remains. \\
        Summary only & Use the noisy scope prediction to identify interference and
        attempt repair, without empirical historical audits. \\
        \rowcolor{ssperowbg}
        SSPE & Use the true anonymous capability context, one risk-directed audit,
        and one random predicted-safe audit when available; repair measured conflicts
        and otherwise reject unsafe proposals. \\
        Full-replay oracle & Check every previous capability and reject any proposal
        whose measured regression remains after repair. This is an upper bound rather
        than a practical baseline. \\
        \bottomrule
    \end{tabular}
\end{table}

The random-replay arm is a lower-audit comparison, not an exactly compute-matched
baseline: it checks one historical capability, whereas SSPE can check two. There is no
separate reject-only arm in the implemented study. These distinctions are important
when interpreting the simulator as mechanism evidence rather than a leaderboard.

\paragraph{Metrics.}

Let $T=Km$ be the number of update steps, $a_t$ the pre-update batch score from
Equation~\ref{eq:synthetic-batch-score}, and
$q_{c}^{\max}=\max_{0\leq t\leq T}q_{t,c}$. We report

\begin{align*}
    A_{\mathrm{preq}} &= \frac{1}{T}\sum_{t=1}^{T}a_t,
    &
    A_{\mathrm{final}} &= \frac{1}{K}\sum_{c=1}^{K}q_{T,c}, \\
    F &= \frac{1}{K}\sum_{c=1}^{K}
    \left[q_c^{\max}-q_{T,c}\right]_+,
    &
    H &= \frac{1}{T}\sum_{t=1}^{T}
    \mathbb{I}\!\left[\exists c\in\mathcal H_t:
    q_{t+1,c}<q_{t,c}\right].
\end{align*}

Here $A_{\mathrm{preq}}$ is cumulative prequential reward,
$A_{\mathrm{final}}$ is final macro-capability performance, $F$ is average
forgetting, and $H$ is the harmful-update rate. We additionally record the no-update
rate, mean immediate current-capability gain, the fraction of that gain retained after
repair, recall of truly harmed historical capabilities, audit executions, and forward
transfer to first-seen capabilities. Audit counts are retained for protocol
transparency and are not treated as a contribution.

Every method receives the same initial state and environmental random draws for a
given configuration and world. Policy-specific random choices use separate seeded
streams so that branching in one policy cannot change another policy's world. Effects
are therefore compared within paired $(\text{configuration},\text{world})$ cells.

\paragraph{Complete controlled results.}

\begin{table}[H]
    \centering
    \small
    \caption{Mean results over the complete 11,300-world sweep. Higher is better for
    cumulative reward and final macro; lower is better for forgetting and harmful
    updates.}
    \label{tab:synthetic-complete-results}
    \begin{tabular}{lrrrr}
        \toprule
        Method & Preq. & Final & Forget. & Harm \\
        \midrule
        Static skill & 0.475 & 0.475 & 0.000 & 0.000 \\
        SkillGrad & 0.520 & 0.462 & 0.387 & 0.514 \\
        SkillOpt & 0.520 & 0.464 & 0.386 & 0.511 \\
        Random replay & 0.553 & 0.570 & 0.271 & 0.355 \\
        Summary only & 0.595 & 0.673 & 0.206 & 0.298 \\
        \rowcolor{ssperowbg}
        SSPE & 0.584 & 0.681 & 0.171 & 0.248 \\
        Full-replay oracle & 0.655 & 0.889 & 0.000 & 0.000 \\
        \bottomrule
    \end{tabular}
\end{table}

On the 7,700 nonlinear paired worlds, SSPE reduces mean forgetting by
$55.84\%$ relative to SkillGrad. Its paired improvements are $+0.3211$ in
final macro-capability performance and $+0.09345$ in cumulative prequential reward,
while its harmful-update rate is $0.3899$ lower. SSPE also has lower forgetting
than random replay. However, immediate current-capability gain is $0.02864$ lower than
SkillGrad. This exceeds the preregistered tolerated loss of $0.02$.
The study therefore supports the interference-repair mechanism while exposing a real
stability--plasticity trade-off. It does not establish superiority to native textual
optimizers or replace the real streaming experiments.

\section{Qualitative Skill and Trajectory Example}
\label{app:qualitative-example}

This section connects the text of an evolved skill to the complete behavior it elicits
on an authentic held-out task. We selected the example \emph{after} observing the
evaluation because it exposes concrete differences among the three methods; it is
therefore illustrative rather than additional quantitative evidence. The task is
\texttt{multi\_turn\_miss\_param\_68} from BFCL's missing-parameter category. It
belongs to the outcome-independent held-out split, was never used for evolution or
verification, and was executed with the same GPT-5.4 model, prompt, tools, and initial
environment for all methods.

\paragraph{Task.}
The task consists of the following six user turns. We reproduce them in full so that
the relationship between the skill instructions and later tool behavior is visible.

\begin{enumerate}
    \item ``I'm about to embark on a road trip adventure and I want my car to be in
    peak condition. Could you make sure to increase the current fuel level to ensure
    that my tank is full, so I don't have to keep stopping to refuel along the way?''

    \item ``Before I hit the open road, I need to get the engine running smoothly.
    Can you confirm there's enough fuel, and ensure the engine's primed for a seamless
    start?''

    \item ``I want to make certain my tires are roadworthy before setting off. If any
    of my car's tires are showing pressure below 40, point me in the direction of the
    closest tire service station, because I definitely don't want to run into tire
    trouble.''

    \item ``Moreover, could you set up the GPS to guide me directly to the nearest
    shop if my tires aren't up to the mark, so I'm not wandering off course?''

    \item ``Once my car is ready for the journey, it would be fantastic to let my
    friend know about my travel plans. Could you draft a tweet that says: `Starting
    my road trip with a car that is fully prepared and raring to go!' with hashtags
    \#Roadtrip \#Adventure and mention him?''

    \item ``His handle is `huanzhimao01'.''
\end{enumerate}

Two dependencies are deliberately revealed only during interaction. On turn 2,
\texttt{startEngine} reports that the doors must be locked and the brake must be
fully pressed. On turns 5--6, the requested post cannot be completed until the user
supplies the missing handle. Success therefore requires both recovery from tool
feedback and retention of an unfinished state-changing action across turns.

\paragraph{Distinguishing parts of the evolved skills.}
The three frozen skills contain broader instructions for agent operation. Below we
reproduce the parts most directly relevant to this task, rather than paraphrasing
them. Their SHA-256 hashes begin \texttt{fda64fe3} (SkillGrad),
\texttt{ed53a4e6} (SkillOpt), and \texttt{09bce676} (SSPE).

\noindent\textbf{SkillGrad.}
\begin{quote}
\small
``For multi-turn tasks, re-evaluate the current-turn objective explicitly after each
user message and after each meaningful tool result. Decide what exact next action,
exact next report, exact bundled report, or exact deliberate non-action that result
authorizes for this turn. If the request implies carrying through a workflow, do not
stop at inspection, diagnosis, recommendation, status reporting, parameter
acknowledgment, or an offer to continue when the needed next action is already
authorized and supported by context.''

``When a tool action fails or reports unmet preconditions, treat that as guidance
about the next required steps rather than as a terminal result. Infer the missing
prerequisites from the observable state, satisfy them if the task authorizes doing
so, and retry the blocked action.''
\end{quote}

\noindent\textbf{SkillOpt.}
\begin{quote}
\small
``Before calling a state-changing tool, check whether the operation has observable
prerequisites or required parameters. If prerequisites are needed and the user's
request implies completing the task end-to-end, satisfy them first when permitted by
the tools rather than attempting the main action prematurely. After intermediate
state changes, use the updated observed state for the next tool call.''

``Before asking the user for more parameters, distinguish between fields explicitly
required by the task or previously observed tool usage and fields you are only
speculating might be needed. Do not block progress by requesting unsupported extra
details. First attempt the smallest valid tool call using arguments grounded in the
user request and prior tool outputs.''
\end{quote}

\noindent\textbf{SSPE.}
\begin{quote}
\small
``On later turns, prefer completing the pending action over giving advisory or
hypothetical text.''

``If the tools can do the requested action, perform it instead of describing how it
could be done.''

``Treat many tool errors as actionable feedback, not dead ends: if starting a vehicle
reports unlocked doors or insufficient brake pressure, satisfy those prerequisites
and retry.''

``Vehicle control: requests to start the engine or navigate may require prerequisite
actions; satisfy obvious prerequisites from tool feedback and continue until the
requested end state is reached or a real blocker remains.''
\end{quote}

All three skills encourage multi-turn continuation. The SSPE artifact additionally
contains task-grounded recovery instructions for the exact class of hidden vehicle
preconditions encountered here, together with an explicit instruction to execute a
pending action rather than merely describe it. We next show how the three frozen
artifacts behave on every turn. Tool calls are rendered as
\texttt{function(argument=value)} for readability while preserving every function
name and argument value.

\paragraph{Complete SkillGrad trajectory.}
\begingroup
\raggedright
\begin{enumerate}
    \item \textbf{Tool calls:}
    \texttt{displayCarStatus(option=fuel)}; then
    \texttt{fillFuelTank(fuelAmount=35)}.

    \textbf{Assistant:} ``Fuel tank filled. Current fuel level: 50.0 gallons.''

    \item \textbf{Tool calls:}
    \texttt{displayCarStatus(option=fuel)}; then
    \texttt{startEngine(ignitionMode=START)}, which reports that the driver,
    passenger, rear-left, and rear-right doors are unlocked;
    \texttt{lockDoors(unlock=false)} for the driver, passenger, rear-left, and
    rear-right doors;
    \texttt{startEngine(ignitionMode=START)}, which reports that the brake must be
    pressed; \texttt{pressBrakePedal(pedalPosition=0.5)};
    \texttt{startEngine(ignitionMode=START)}, which reports that the brake must be
    pressed fully; \texttt{pressBrakePedal(pedalPosition=1)}; and finally
    \texttt{startEngine(ignitionMode=START)}, which succeeds.

    \textbf{Assistant:} ``Confirmed: fuel is sufficient at 50.0 gallons, and the
    engine is now running smoothly. Also completed the start prerequisites: Doors
    locked; Brake fully pressed; Engine started successfully. Battery voltage:
    12.6 V.''

    \item \textbf{Tool calls:} \texttt{check\_tire\_pressure()}; then
    \texttt{find\_nearest\_tire\_shop()}.

    \textbf{Assistant:} ``All four tires are below 40 psi: Front left: 35.0 psi;
    Front right: 35.0 psi; Rear left: 35.0 psi; Rear right: 35.0 psi. Nearest tire
    service station: 456 Oakwood Avenue, Rivermist, 83214.''

    \item \textbf{Tool call:}
    \texttt{set\_navigation(destination=456 Oakwood Avenue, Rivermist, 83214)}.

    \textbf{Assistant:} ``GPS is set. Now navigating to 456 Oakwood Avenue,
    Rivermist, 83214.''

    \item \textbf{Tool calls:} none.

    \textbf{Assistant:} ``Starting my road trip with a car that is fully prepared
    and raring to go! \#Roadtrip \#Adventure @yourfriend''

    \item \textbf{Tool calls:} none.

    \textbf{Assistant:} ``Starting my road trip with a car that is fully prepared
    and raring to go! \#Roadtrip \#Adventure @huanzhimao01''
\end{enumerate}
\endgroup

The official evaluator credits the first five turns but fails the task because the
sixth turn contains no \texttt{post\_tweet} call. SkillGrad obtains a $5/6$ correct
turn prefix.

\paragraph{Complete SkillOpt trajectory.}
\begingroup
\raggedright
\begin{enumerate}
    \item \textbf{Tool calls:}
    \texttt{displayCarStatus(option=fuel)}; then
    \texttt{fillFuelTank(fuelAmount=35)}.

    \textbf{Assistant:} ``Fuel tank topped off to full. Current fuel level:
    50.0 gallons.''

    \item \textbf{Tool calls:}
    \texttt{displayCarStatus(option=fuel)} and
    \texttt{displayCarStatus(option=engine)}; then
    \texttt{startEngine(ignitionMode=START)}, which reports that all four doors are
    unlocked.

    \textbf{Assistant:} ``There's enough fuel: 50.0 gallons. I tried to start the
    engine, but it couldn't start because all four doors are currently unlocked:
    driver, passenger, rear\_left, rear\_right. If you want, I can lock all doors and
    start the engine next.''

    \item \textbf{Tool calls:} \texttt{check\_tire\_pressure()}; then
    \texttt{find\_nearest\_tire\_shop()}.

    \textbf{Assistant:} ``All four tires are below 40 psi: Front left: 35.0 psi;
    Front right: 35.0 psi; Rear left: 35.0 psi; Rear right: 35.0 psi. Nearest tire
    service station: 456 Oakwood Avenue, Rivermist, 83214.''

    \item \textbf{Tool call:}
    \texttt{set\_navigation(destination=456 Oakwood Avenue, Rivermist, 83214)}.

    \textbf{Assistant:} ``GPS is set. Destination: 456 Oakwood Avenue, Rivermist,
    83214.''

    \item \textbf{Tool calls:} none.

    \textbf{Assistant:} ``I can draft it, but to actually include the mention I need
    your friend's Twitter handle. Draft: Starting my road trip with a car that is
    fully prepared and raring to go! \#Roadtrip \#Adventure. Send me his @username
    and I'll include it.''

    \item \textbf{Tool calls:} none.

    \textbf{Assistant:} ``Starting my road trip with a car that is fully prepared
    and raring to go! @huanzhimao01 \#Roadtrip \#Adventure''
\end{enumerate}
\endgroup

The official evaluator reports an incorrect environment state: the engine remains
stopped, the doors remain unlocked, and the brake remains released. It also observes
no tweet-posting side effect. SkillOpt obtains a $1/6$ correct turn prefix.

\paragraph{Complete SSPE trajectory.}
\begingroup
\raggedright
\begin{enumerate}
    \item \textbf{Tool calls:}
    \texttt{displayCarStatus(option=fuel)}; then
    \texttt{fillFuelTank(fuelAmount=35)}.

    \textbf{Assistant:} ``Fuel tank is now full at 50.0 gallons.''

    \item \textbf{Tool calls:}
    \texttt{displayCarStatus(option=fuel)} and
    \texttt{displayCarStatus(option=engine)}; then
    \texttt{startEngine(ignitionMode=START)}, which reports the unlocked doors;
    \texttt{lockDoors(unlock=false)} for the driver, passenger, rear-left, and
    rear-right doors;
    \texttt{startEngine(ignitionMode=START)}, which reports the missing brake
    pressure; \texttt{pressBrakePedal(pedalPosition=0.5)};
    \texttt{startEngine(ignitionMode=START)}, which reports that the brake must be
    pressed fully; \texttt{pressBrakePedal(pedalPosition=1)}; and finally
    \texttt{startEngine(ignitionMode=START)}, which succeeds.

    \textbf{Assistant:} ``Fuel confirmed: 50.0 gallons. Engine is now running.''

    \item \textbf{Tool calls:} \texttt{check\_tire\_pressure()}; then
    \texttt{find\_nearest\_tire\_shop()}.

    \textbf{Assistant:} ``456 Oakwood Avenue, Rivermist, 83214''

    \item \textbf{Tool call:}
    \texttt{set\_navigation(destination=456 Oakwood Avenue, Rivermist, 83214)}.

    \textbf{Assistant:} ``Navigating to 456 Oakwood Avenue, Rivermist, 83214''

    \item \textbf{Tool calls:} none.

    \textbf{Assistant:} ``Who should I mention in the tweet?''

    \item \textbf{Tool call:} \texttt{post\_tweet} with content
    ``Starting my road trip with a car that is fully prepared and raring to go!'',
    tags \texttt{[\#Roadtrip, \#Adventure]}, and mention
    \texttt{[@huanzhimao01]}.

    \textbf{Assistant:} ``Tweet posted.''
\end{enumerate}
\endgroup

The official evaluator records no error and credits all six turns. SSPE reaches the
requested vehicle state, preserves the navigation state, and performs the final
external posting action, obtaining a $6/6$ correct turn prefix.

\paragraph{Interpretation.}
The example separates skill text from downstream execution. All three artifacts
contain broadly sensible instructions, yet their frozen executions differ at two
points that require procedural continuation. SkillOpt reports the first recoverable
engine precondition instead of satisfying it. SkillGrad recovers from the vehicle
errors but renders the final tweet as text rather than performing the requested side
effect. SSPE both follows the error-revealed prerequisite chain and converts the
later handle into the pending \texttt{post\_tweet} call. Its concrete clauses align
with these behaviors, but this post-hoc case is not a causal comparison by itself;
the aggregate held-out evaluation remains the primary evidence.

\end{document}